# Testing Causal Explanations: A Case Study for Understanding the Effect of Interventions on Chronic Kidney Disease

Panayiotis Petousis[1], PhD; David Gordon[2]; Susanne B. Nicholas[3], MD, MPH, PhD; and Alex A. T. Bui[1,2], PhD, on behalf of CURE-CKD

[1]Clinical and Translational Science Institute, University of California, Los Angeles Los Angeles, CA

[2]Medical & Imaging Informatics Group University of California, Los Angeles, Los Angeles, CA

[3]Department of Medicine, Division of Nephrology University of California, Los Angeles, Los Angeles, CA

## Abstract

Randomized controlled trials (RCTs) are the standard for evaluating the effectiveness of clinical interventions. To address the limitations of RCTs on real-world populations, we developed a methodology that uses a large observational electronic health record (EHR) dataset. Principles of regression discontinuity (rd) were used to derive randomized data subsets to test expert-driven interventions using dynamic Bayesian Networks (DBNs) do-operations. This combined method was applied to a chronic kidney disease (CKD) cohort of more than two million individuals and used to understand the associational and causal relationships of CKD variables with respect to a surrogate outcome of ≥40% decline in estimated glomerular filtration rate (eGFR). The associational and causal analyses depicted similar findings across DBNs from two independent healthcare systems. The associational analysis showed that the most influential variables were eGFR, urine albumin-to-creatinine ratio, and pulse pressure, whereas the causal analysis showed eGFR as the most influential variable, followed by modifiable factors such as medications that may impact kidney function over time. This methodology demonstrates how real-world EHR data can be used to provide population-level insights to inform improved healthcare delivery.

## 1. Introduction

Randomized controlled trials (RCTs) are the *de facto* gold standard for evaluating the effectiveness of a given intervention, providing insights into treatment effects, and, ultimately, offering explanatory causal relationships [1][2]. By their nature, RCTs can have limitations when generalizing to broader populations (e.g., due to strict inclusion and exclusion criteria [3]), and their conduct can be costly, time-consuming, and difficult given pragmatic, ethical, legal, and social concerns [4]. As such, alternative methods have been established and adopted over many years to provide (statistical) evidence informing the reasons for observed effects and to strengthen a belief in causal connections around a given treatment. Well-known techniques – which all approach the problems in different ways – include propensity score matching [4], instrumental variable analysis [5], Cox proportional hazards [6], and causal inference methods [7]. While these methods are well-understood and commonly used, they can suffer from a variety of shortcomings, such as overfitting [4], indirect causal interpretations [6], and high computational complexity [7].

To address these limitations, we explored a new methodology for emulating an RCT from observational data. Using a dynamic Bayesian network (DBN) to model a target outcome of interest over time, we adapted principles from regression discontinuity and do-calculus to derive an algorithm that generates randomized datasets and tests specific interventions using do-operations [8,9]. This combined framework, called *rd-do*, allows us to simulate different situations to ascertain the effect of a given variable (modeled by the DBN) in the absence of confounders through their explicit removal. To test this approach, we employed a large dataset drawn from the electronic health records (EHRs) of more than two million individuals as part of a prospective chronic kidney disease (CKD) registry. From this data, DBNs were developed to predict if an individual would experience a ≥40% decline in estimated glomerular filtration rate (eGFR) in future years [10]. Using the rd-do algorithm on DBNs from two large health systems, we demonstrate the average causal effect of CKD variables on the outcome of ≥40% eGFR decline. Our analysis suggests the potential of this approach in providing causal evidence. This paper details the algorithm's components and evaluation and discusses rd-do as a path toward causal interpretation.

## 2. Materials and Methods

### 2.1 Motivating background

Several statistical methods have been used to simulate RCT analyses. Propensity score matching focuses on identifying similar groups to estimate the effect of treatment while accounting for covariates that predict administering treatment. The correct estimate of the effect of treatment depends on the generation of matched randomized control and treatment groups, which use the score of a model (e.g., logistic regression) to match patients. However, this method is prone to overfitting (e.g., many predictors related to the number of events) and stratification limitations when the number of covariates is large [4]. In comparison, in Cox proportional hazard models, even though suitable for large datasets, the hazard ratio cannot be interpreted as causal [6]. Instrumental variable analysis, a technique often used by economists to estimate the effects of treatment variables, can suffer from the choice of a bad instrument (i.e., one correlated with omitted variables) [5] and is less suited to the continuously evolving healthcare setting in which new variables are introduced (e.g., use of a new drug or protocol). Lastly, causal inference methods involve using a graphical model to identify the causal effect of variables within or without a given dataset [11]. Using acyclic graphical models as carriers of independence assumptions has been essential in modeling interventions, and one can quantify interventions and counterfactual reasoning by utilizing the calculus of interventions [11]. Nevertheless, identifying the correct causal model is a time-consuming and computationally expensive task and a function of population and domain knowledge. Using a biased population for causal inference can result in a skewed distribution of the average causal effect [12]. Additionally, given the complexity of the model, some computations can become intractable when applied to a large population. As such, obtaining a sufficiently large and randomized dataset in combination with a causal model is requisite to obtain accurate population-level causal effects.

## 2.2 The rd-do algorithm

To overcome the potential issues described, we combine two approaches in our rd-do algorithm to operate on DBNs and test different interventions: we applied regression discontinuity to identify a randomized subset of our dataset around a given model threshold (cut point) and then use the pretrained DBN to execute do()-based interventions to estimate the average causal effect of variables on the outcome when a causal path exists.

### 2.2.1 *Regression discontinuity*

Regression discontinuity (RD) is a way of estimating treatment effects in a nonexperimental way [8]. The main idea of RD is that individuals with scores around a given decision cutoff point represent a randomized sample of the population, given an assumption that there is imprecise control over the assignment variable [8]. To test that a randomized sample exists around the cutoff, baseline characteristics – that is, variables determined prior to the realization of the assignment variable — should have the same distribution above and below the cutoff [8]. Figure 1 depicts the sample space around a given threshold. Notably, contemporary views describe RD as its own type of study design that most closely resembles a randomized experiment [8]. Here, we employed an approach to randomizing control and treatment groups via sharp regression discontinuity design (RDD). Sharp RDD is known as a special case of matching [13] in contrast to fuzzy RDD, which is a type of instrumental variable regression [14]. Sharp RDD assumes the probability of treatment jumps from zero to one at the threshold, which we used in our implementation [15].

### 2.2.1 *Do-operations.*

Do-calculus, which is described for Bayesian (belief) networks, can provide estimates of the causal effect of variables based on the structure of a directed acyclic graph (DAG) and the outcome variable [16]. The classic do($X=x$) operation represents an intervention on an observed or modifiable variable, which can be interpreted twofold: 1) an intervention on a variable $X$ that was observed and thus may have been confounded by other, possibly unknown variables and 2) an intervention by an action (e.g., the use of medication) that is free of confounders. In this light, observing a variable would be incorrect in the presence of unknown or known confounders (i.e., $P(Y \mid X) \neq P(Y \mid do(X))$). Per this theory, to remove the effect of confounders and estimate the causal effect of do($x$) [9], one must delete incoming arrows into variable $X$ and the corresponding equations from the functional causal model. In practice, we need to delete all incoming arrows of a variable $X$ (i.e., transform it into a root variable) and instantiate $X$ to an observation $x$; existing software packages like SamIam [17] and GeNie [18] can then estimate the causal effect on this mutilated graph by estimating the posterior (Figure 2). We used this strategy to instantiate $X$ with an observation $x$ and estimate a causal effect.

Table 1 summarizes a description of the rd-do algorithm. Conceptually, the first step involves using an optimal threshold in the probability space for the given target outcome. Around that threshold, we next use a sampling window to select a different number of samples (Figure 1) and tested for statistical significance between baseline covariates to the left and right of the threshold. If none are found to be statistically significant and thus randomized, we then apply do() operations for all variables of interest with respect to the outcome. The causal model prediction for each variable $X$ observation, $x_i$, is computed for each sample and used to compute the average causal effect.

2.3 Experimental setup

To test the rd-do algorithm, we used a data-driven DBN created in ongoing work toward modeling CKD-related health outcomes [19]. The DBN focused on predicting individuals who will experience a future ≥40% eGFR decline. We describe the dataset and its use to construct the model and its validation to contextualize how it was used to assess the rd-do algorithm:

*2.3.1 Dataset and population definition.*

We used the EHRs from individuals with or deemed at-risk for the development of CKD from the Center for Kidney Disease Research, Education, and Hope (CURE-CKD) Registry [20]. In this registry, individuals were at-risk for CKD were comprised of those with an existing diagnosis of prediabetes, diabetes, hypertension, and/or eGFR <60 mL/min/1.73 m². Data covered individuals from 1/1/06-12/31/20 (n=2,250,806) from two non-profit healthcare systems, Providence Health System (PHS, n=1,917,619) and the University of California at Los Angeles Health (UCLA, n=333,187). Harmonized data from these institutions includes information on demographics, vitals, laboratory measurements, prescribed medications, and diagnostic and procedure codes. **Figure 3** illustrates key time points of data collection as part of the CURE-CKD Registry: 1) study entry, representing the first day of inclusion into the registry, at which point multiple baseline observations are recorded; 2) 90-days post-entry, during which additional clinical observations are captured; and 3) follow-up, in which all variables are monitored annually. **Figure 4** depicts the STROBE diagram that defines the dataset for creating the DBN with additional inclusion criteria. Additional details on the inclusion criteria can be found in **Tables 2 and 3**, which also include descriptive statistics for the variables used in the DBN and, thus, our rd-do evaluation.

*2.3.2 Data preprocessing and dataset preparation.*

Based on input from expert nephrologists, outlier values were set to missing values. We employed a minimum description length (MDL) discretization via the Orange package [21] to create discrete bins automatically on continuous variables. Three dataset groupings were created: disaggregated by site (i.e., UCLA, PHS) and a combined dataset (i.e., UCLA + PHS). Each of these datasets was further stratified, based on the outcome of ≥40% eGFR decline at least once during the follow-up period, and then finally randomly split into training (60%), validation (20%), and testing sets (20%). Stratification was based on maintaining the same proportion of positive cases (≥40% eGFR decline) across each group.

*2.3.3 DBN learning and evaluation.*

Model topology was learned automatically using our Ranking Approaches for Unknown Structures (RAUS) framework [22,23]. Three models were learned (Figures 5-7): the UCLA+PHS, PHS, and UCLA models. The learned DBNs were parameterized using the expectation maximization (EM) algorithm [24]. Model training was performed on a random under-sampled subset of the training set to maintain a 1:1 ratio of ≥40% eGFR decline cases to non-decliners. All site models' results are described in [19], and all models[1] and code[2] are publicly available. Briefly, in evaluation, DBN performance improves significantly over time in terms of discrimination and precision in finding ≥40% eGFR decline cases. All models' evaluation metrics over time are publicly available[3]. Of relevance to this rd-do study is to identify an optimal threshold for classifying a patient as ≥40% eGFR decline [19], we used the Youden statistic [25]. Given these DBNs, we performed two analyses with rd-do using the hold-out test set for each site from our CURE-CKD Registry:

1. *Associational exploratory analysis.* Using the rd-do algorithm but without a corresponding do() operation (i.e., only computing the observational probability of observing a certain variable), we computed the associational effect of each variable used in our study. This aspect of the analysis intends to explicate associations observed in the literature around CKD.
2. *Causal explanatory analysis.* Using the rd-do algorithm, we computed the interventional probability for each variable and subject in our randomized dataset based on a causal pathway from the variable to the outcome. The results were averaged over individuals to compute the average causal effect of each variable's category. From this analysis, we will understand which variables have a statistically significant causal effect and compare it with their associational effect.

To set an RD design, we used a $\chi^2$ test to test the null hypothesis that the distributions of study entry covariates for patients to the right and left side of the threshold are similar. Specifically tested covariates included age, race, ethnicity, eGFR, CKD status, and diabetes diagnosis at study entry. Bonferroni corrections were applied for each statistical test. The RD design was set at each time point of observing the outcome of ≥40% eGFR decline (i.e., a new randomized sample around each threshold). For each randomized threshold for our variable of interest, we estimated the associational ($Pr(Y|X=x_i)$) or causal probability ($Pr(Y|do(X=x_i))$) by setting each category, $x_i$, as evidence. Each category can be considered an arm of a clinical trial using the same randomized sample. The probability distribution collected for each category, $x_i$, was assumed to be derived from independent samples as the value for each sample's X variable was set to the category $x_i$, thus synthesizing two independent samples for a pair of categories. The probability distributions associational ($Pr(Y|X=x_i)$) or causal probability ($Pr(Y|do(X=x_i))$) of each category $x_i$ were tested using the two-sample Kolmogorov-Smirnoff test (p-value <0.05 as significant) with the null hypothesis that they are drawn from the same continuous distribution. Rejecting the null hypothesis would indicate a difference in the associational/causal effect in a pair of $x_i$ categories. Subsequently, variables with the highest statistically significant average association/causal effect difference between a pair of categories were ranked first in importance and explained in our Results section. This was performed for all variables prior to the time point of the outcome of interest in the DBN model.

## 3. Evaluation and Results

Three models were used over six annual time points of ≥40% eGFR decline to generate associational and causal effects using rd-do. The time complexity of the rd-do algorithm is O(n), with n depending on three parameters: the number of windows, the number of data points in the best window, or the number of covariates assessed. These experiments were conducted on an Ubuntu 18.04.6, 64-bit, Intel® Core™ i9-9900 CPU @ 3.10GHz × 16 with 126 GB of RAM machine using Python 3.7.4.

---

[1] https://zenodo.org/uploads/13787736

[2] https://github.com/uclamii/DBN_simulation_modeling

[3] https://zenodo.org/uploads/13787768

## 3.1 Associational explanatory analysis

Figure 8 depicts statistically significant association effects ($p<0.05$) for each model (UCLA, PHS, UCLA+PHS), ranked by effect size. Tighter error bars (standard deviation) represent a more direct and stronger effect of each variable's category on the outcome. We found similarity across the sites for variables with the highest associational effect differences. As anticipated, as eGFR increases, the average associational effect of a ≥40% decline in eGFR decreases. Specifically, when eGFR values exceed 60 mL/min/1.73 m², the average effect falls below the threshold needed to classify an individual as experiencing a ≥40% decline in the following year. We also note that the longer use of proton pump inhibitors (PPIs) the year before observing a ≥40% eGFR decline increased the average associational effect of the ≥40% eGFR decline. Urine albumin-to-creatinine ratio (UACR) and pulse pressure (PP) in the year preceding ≥40% eGFR decline have a monotonically increasing trend with respect to the average associational effect.

## 3.2 Causal explanatory analysis

Figure 9 presents statistically significant average causal effects ($p<0.05$) of each variable's category (e.g., for the eGFR variable, a category represents a bin with a range of 0-15 mL/min/1.73 m²) with respect to the target outcome. Again, we observed similarities in the findings across the models and sites. In the underlying DBNs, the eGFR variable is directly connected with the target outcome, we expectedly see the highest average causal effects with tight error bounds as eGFR moves outside the CKD range (0-60 mL/min/1.73 m²). Beyond this lab variable, significant causal effects were also found with certain medications: angiotensin-converting enzyme (ACE) inhibitors and angiotensin receptor blockers (ARBs), used to lower blood pressure, had, on average, a decreasing trend of the average causal effect (probability of ≥40% eGFR decline in the next year) with longer coverage periods. Interestingly, evaluation of ACE inhibitor/ARB usage using the PHS model on the UCLA population showed a significant decrease in the average causal effect from Years 4-5. The causal relationship of PPIs, two years preceding the outcome, was also noted but was mediated through intermediate variables (mediators) [16] per the DBN topology. High UACR, low eGFR, and medication use resulted in larger error bars, as fewer patients with these values were observed compared to the healthy population. Moreover, patients with these data typically had more severe disease(s) status with complications.

## 4. Discussion

A ≥40% decline in eGFR is a significant clinical endpoint in the progression of CKD and can predict progression to kidney failure within two years [10,26]. Importantly, there is an increasing number of individuals at risk of developing both CKD and kidney failure [27,28] [29–34]. Thus, predicting and slowing eGFR decline is essential. In this light, we leveraged EHR-based data to test a novel algorithm, rd-do, to simulate a clinical trial and provide insights into modifiable factors in CKD patients.

Our methodology was built atop regression discontinuity and the ability of belief networks for causal inference. We generated a random sample around the threshold of the target outcome (≥40% eGFR decline) and used this sample with a DBN model to simulate the individualized causal effect of a variable on the outcome of interest by modifying the values of certain variables via do() operations. Using do() operations on random samples provided a way to handle (hidden) confounders, creating a perspective analogous to an RCT by calculating an average causal treatment effect. Removing the do() operation from our technique is akin to estimating an average associational effect.

Using rd-do, we found that the average associational effect for each DBN (UCLA, PHS, UCLA+PHS)

exhibited similar results, with the most influential variables to the outcome being eGFR, UACR, and pulse pressure. PPIs were the most influential medication on the outcome, with longer usage increasing the likelihood of ≥40% eGFR decline. These findings mirror other studies in the clinical literature [29–32]. Our causal analysis showed eGFR as the most influential variable, followed by modifiable factors such as medications [41]. The causal analysis consistently showed a reduction of ≥40% eGFR decline risk given an increase in the number of days of ACEi/ARB prescription, again a finding consistent with the clinical literature and current guideline-directed medical therapies [33–35]. Using a causal effect analysis, potential trends of certain variables, such as pulse pressure, became consistent compared to the associational trends. In the causal analysis, PP showed a consistent reduction of ≥40% eGFR decline risk, given an increasing PP measurement. However, the effect of PP is mediated through multiple medications.

One advantage of our method with respect to a clinical trial is the use of the same random sample in the control and treatment arm, eliminating the problems around selecting a randomized matched control group. Another potential strength of rd-do is the use of belief networks, which can help overcome issues related to the presence of missing values typical of EHRs and the myriad reasons for such missingness (e.g., missing not at random, such as a patient lost to follow-up; missing at random, for incomplete surveys or records; irregular frequency of lab observations, etc.) [36–38]. Nevertheless, the rd-do algorithm has several limitations: 1) the chosen threshold might not be optimal, or there might exist multiple optimal thresholds; 2) the window size used might be narrow and yield less precise estimates; and 3) the DBN topology might be incomplete or inaccurate.

Selecting an optimal intervention threshold depends on several factors, some of which are secondary but are related to the outcome of interest. Competing risk factors, such as heart failure or social determinants of health, which may not be part of our registry, can contribute to a patient's adverse outcome. Therefore, intervening and prescribing medication or identifying a patient at risk for a ≥40% eGFR decline can be due to non-kidney disease factors affecting kidney function. Our technique of regression discontinuity for selecting random samples can be extended to use multiple important thresholds across the probability space and assess if the observed patterns are consistent or specific for certain patient profiles.

The window of regression discontinuity determines the size of the random sample used to estimate causal and associational effects. In this study, we chose the smallest random subset (i.e., window) of data points with the highest power to improve computational efficiency. However, further investigation needs to be performed to assess if equivalent larger random subsets with the same power result in the same associational or causal patterns. Moreover, some statistical tests are prone to sample size, as the p-value depends on the sample size.

Interestingly, the DBN structures derived from two different site populations, and their combination showed a similar topology and similar performance metrics when the two populations were evaluated with each model. We observed similar associational and causal effects across the two sites. Nevertheless, an analysis using other methods for learning DBN topologies should be assessed and compared with expert-defined topologies.

A common disadvantage of EHR observational studies is the biased data generation process. For example, patients with severe disease status are more likely to have more observation points than their healthy counterparts, altering the perception of the natural disease progression. We tried to address the bias by using a large causal network (e.g., pretrained DBNs) trained on the training set rather than the selected random sample of records using RD.

Future work will improve insight generation, including studying optimal thresholds and window sizes in the probability space. We performed the causal analysis on a univariate basis; our next steps will involve using multivariate analysis to further study interactions and causal effects.

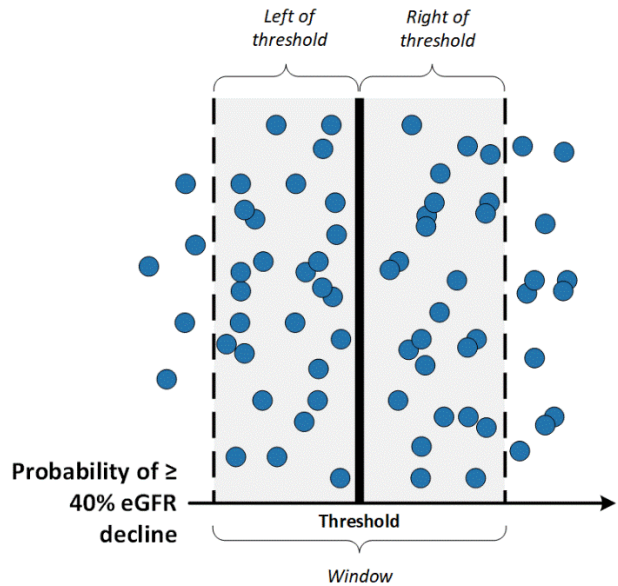

**Figure 1.** RD design, setting window around the threshold and identifying a sample around the threshold.

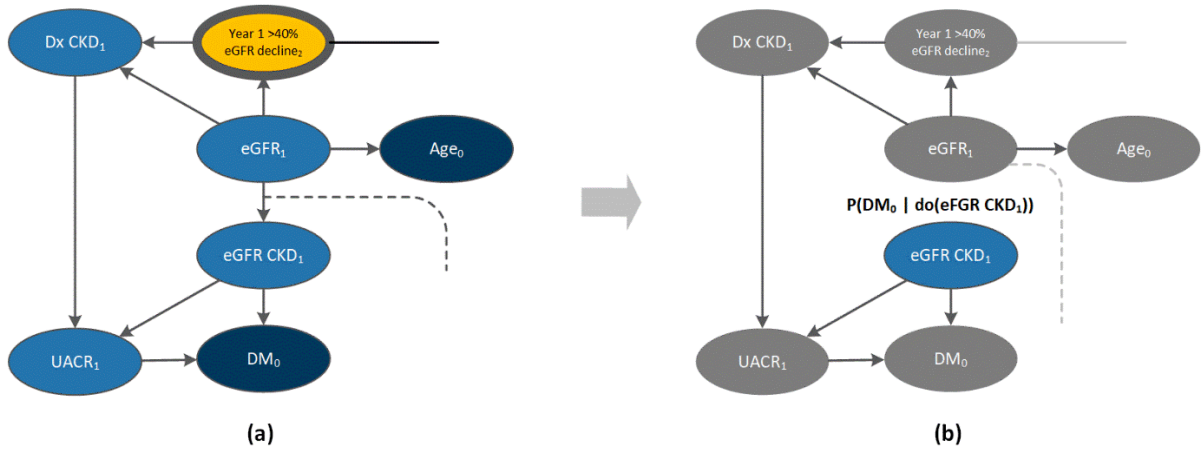

**Figure 2.** (a) UCLA+PHS blanket of dynamic Bayesian network using seven variables. (b) Mutilated DBN through removal of incoming arrows in the variable that do() operation is performed.

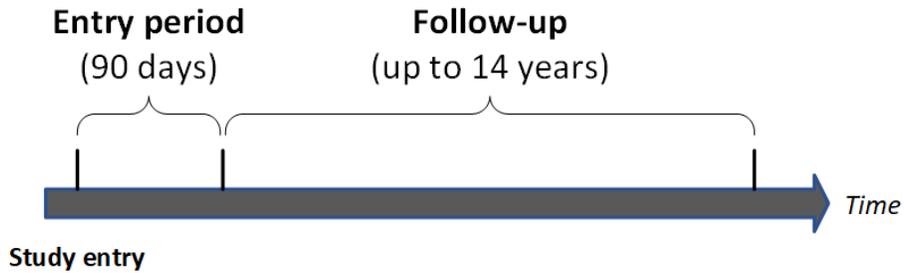

**Figure 3.** CURE-CKD data generation timeline. Study entry represents the day of entering the study ($t_0$). The entry period is the first 90 days after study entry ($t_1$). The follow-up period is a 14-year period ($t_2$- $t_n$). Adopted from [19].

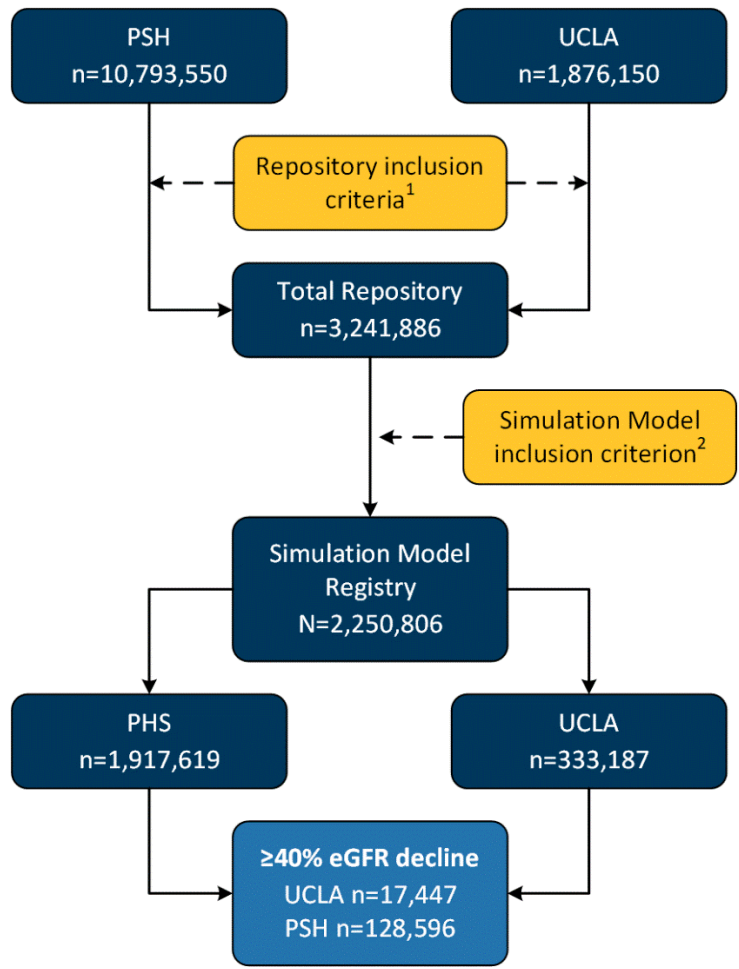

**Figure 4.** STROBE diagram. Adopted from [19].

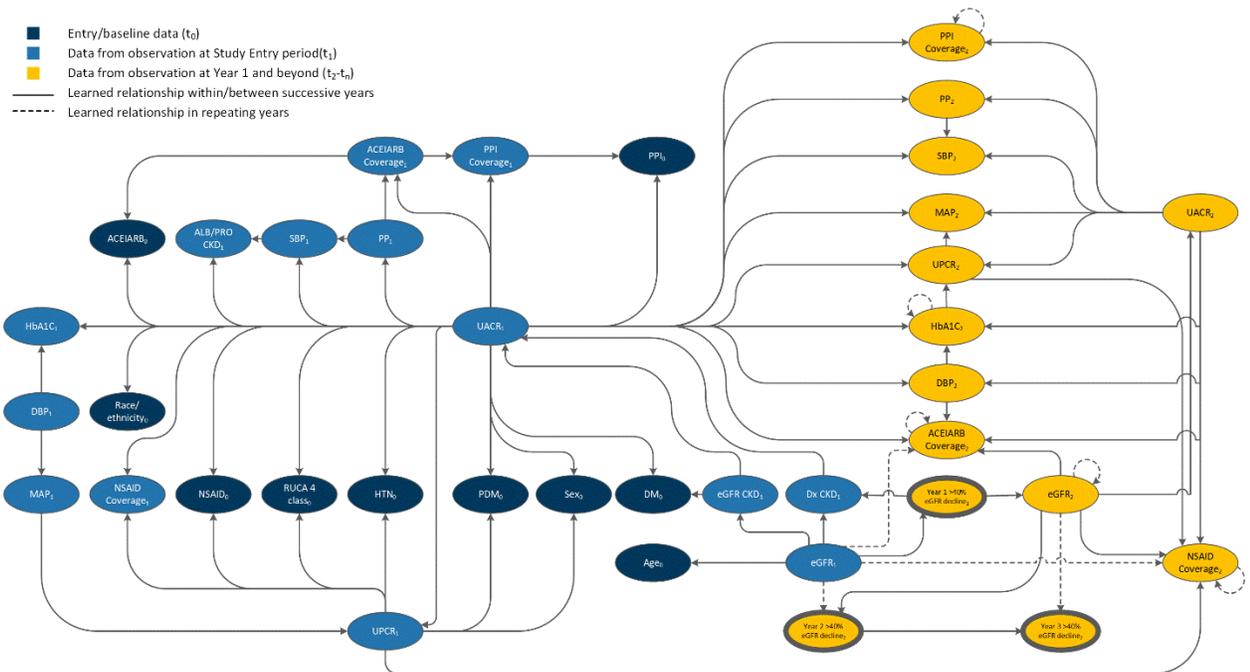

**Figure 5.** UCLA-PHS dynamic Bayesian network. Solid arrows represent statistical dependence and dashed arrows represent temporal statistical dependence. Adopted from [19].

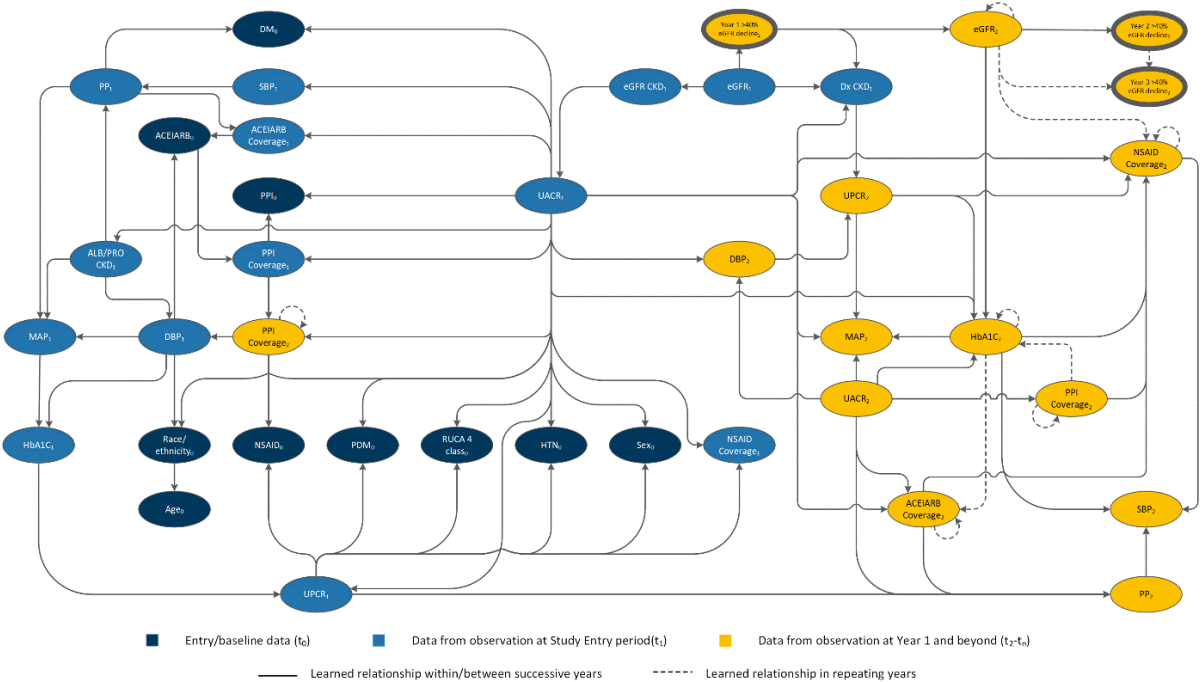

**Figure 6.** PHS dynamic Bayesian network. Solid arrows represent statistical dependence and dashed arrows represent temporal statistical dependence. Adopted from [19].

**Figure 7.** UCLA dynamic Bayesian network. Solid arrows represent statistical dependence and dashed arrows represent temporal statistical dependence. Adopted from [19].

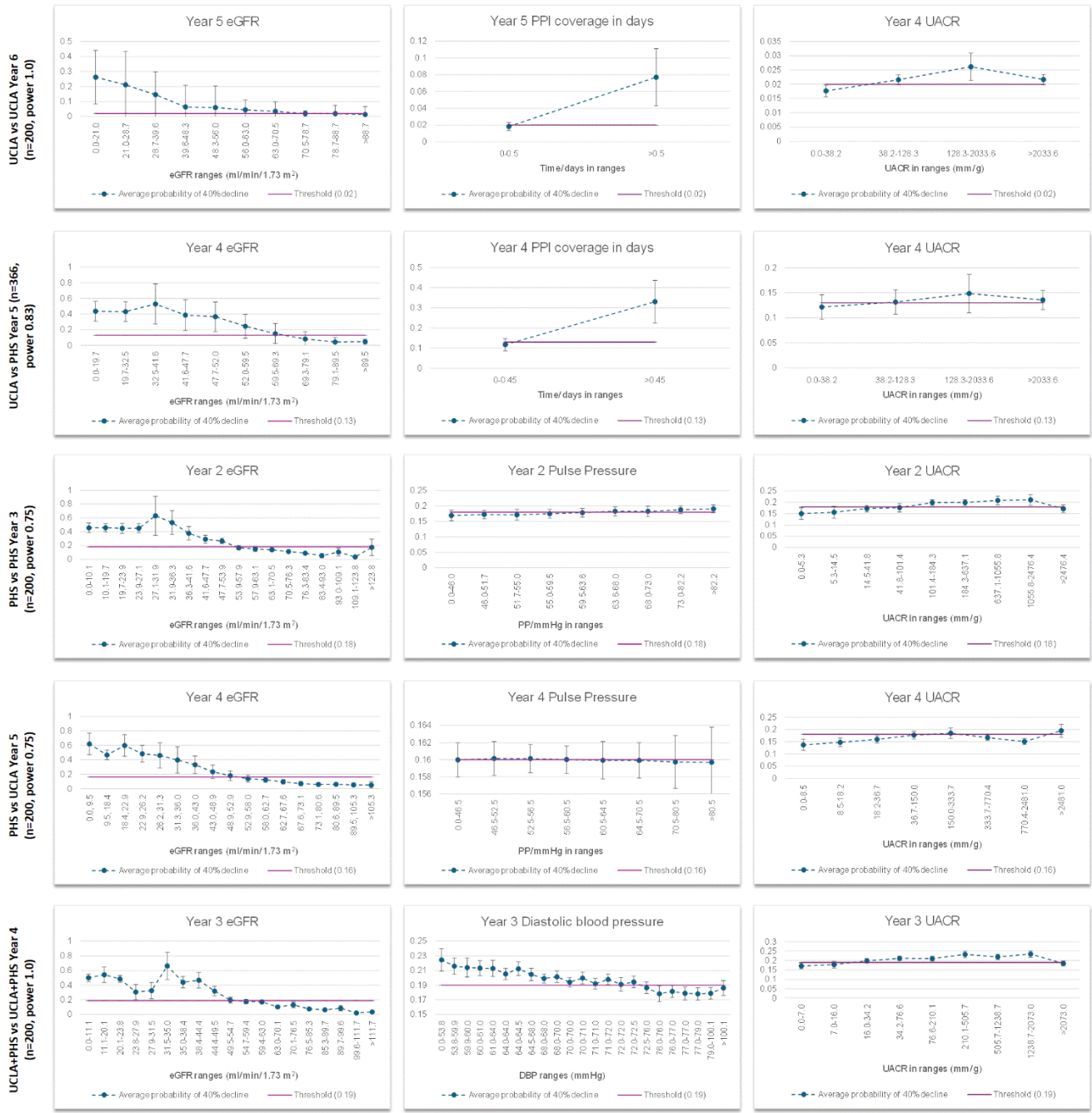

**Figure 8.** Average associational effect of variables with the outcome of 40% eGFR decline. Each model is tested on each site, and the outcome of a 40% eGFR decline with the highest power is shown across the 6-year horizon. The x-axis represents the discretized bins of each variable. Error bars represent the standard deviation for each bin's average probability of 40% eGFR decline.

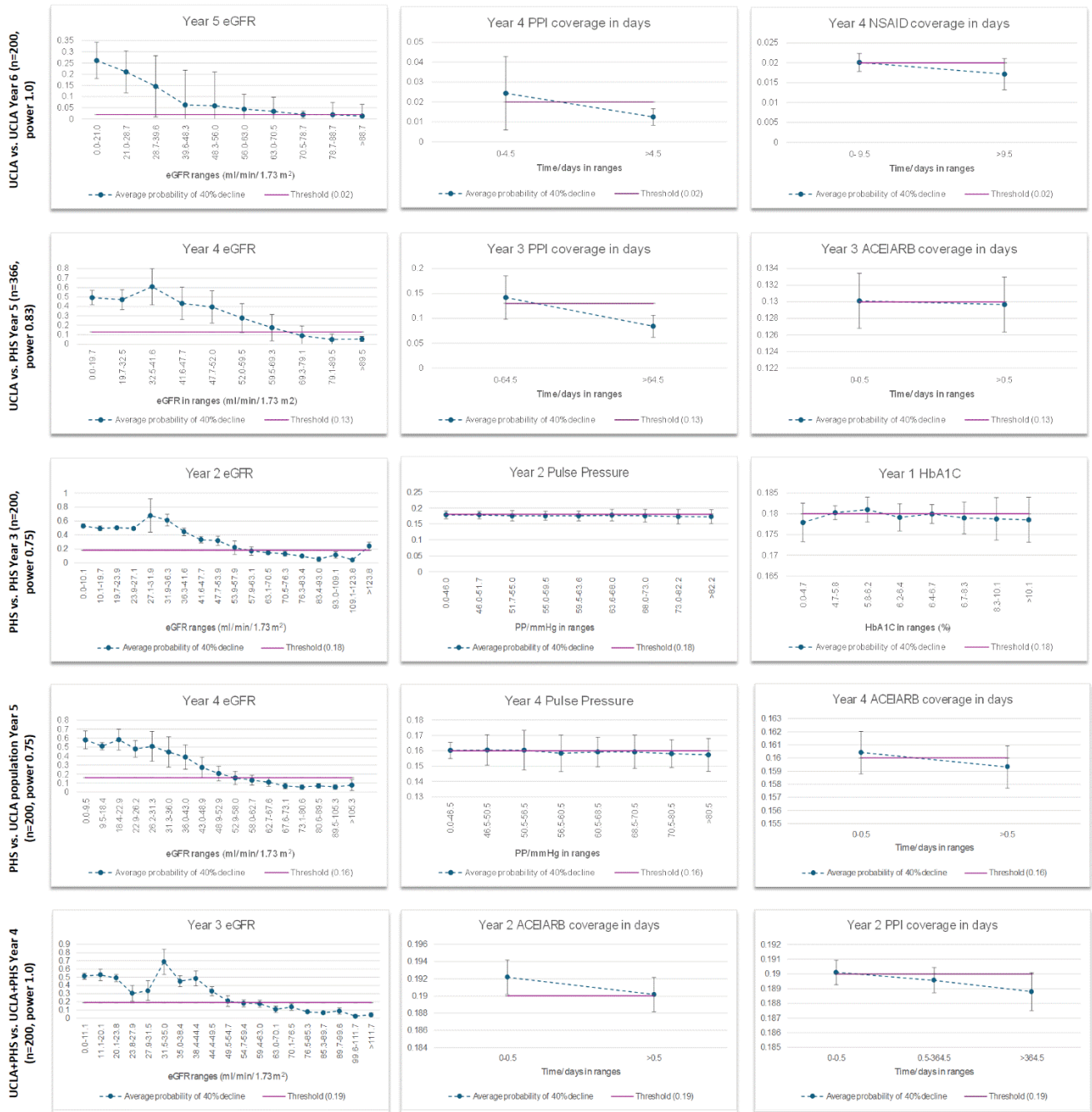

**Figure 9.** Average causal effect of variables with outcome of 40% eGFR decline. Each model is tested on each site, the outcome of 40% eGFR decline with highest power is shown across the 6-year horizon. x-axis represents discretized bins of each variable. Error bars represent the standard deviation for the average probability of 40% eGFR decline in each bin.

**Table 1.** rd-do algorithm for an average causal effect.

**Input**   Test set with covariates and model predictions, set of variables to perform do() operations, baseline covariates, optimal probability threshold, DBN model

**Output**  Table with average causal effect

1. At the probability threshold
2. For each *k*, in a range of *i* to *N* in steps of 1 (with i=200)
3. Define a window around the threshold based on the total number of samples (must match the number of Step 2) to the left and right of the threshold. A window represents the first *k* samples with the minimum absolute distance from the threshold
4. Using baseline covariates, test the null hypothesis if the baseline covariates distribution of the population left of the threshold and the population to the right of the threshold are similar using the chi-squared test
5. Compute the sample power = 1-FP/(FN+FP)
6. If at least one baseline covariate test rejects the null hypothesis
7.    This is not a random sample
8. Else
9.    It is a random sample,
10. Pick the first window with the highest power that corresponds to a random sample
11. If there exists a window that fulfills Step 10
12.    Use pre-trained DBN and the window sample to a) Define the do(X=x) operation; b) Use the DBN to create a causal DBN (i.e., mutilate DBN)
13.    For each category $x_i$
14.       For every data point in the window population
15.          Perform individualized do(X= $x_i$) for causal estimate or without do(X= $x_i$) estimate associational probability
16.          Estimate P(Y|Do(X= $x_i$), Z) while instantiating individualized observations for the remaining variables OR Estimate P(Y| X= $x_i$, Z) while instantiating individualized observations for the remaining variables
17.          Estimate the average causal effect of X= $x_i$ OR Estimate the average associative effect of X= $x_i$
18.          Use the Kolmogorov-Smirnoff test to test the null hypothesis if the underlying distributions of effects for each $x_i$ are statistically significantly different
19.          If significant
20.             Plot the average causal or associational effect using standard deviation as bounds for each category

**Table 2.** Categorical variables, proportions, and percentage of missing data. Adopted from [19].

| Variables | Categories | Proportions (%) | Missing values (%) |
|---|---|---|---|
| Site | UCLA | 14.80 | 0.00 |
| | PHS | 85.20 | |
| Sex | Male | 45.44 | 0.00 |
| | Female | 54.56 | |
| Race/Ethnicity | White | 68.47 | 0.00 |
| | Hispanic | 3.63 | |
| | Black | 4.66 | |
| | Asian | 6.33 | |
| | Native American | 0.95 | |
| | Hawaiian | 0.60 | |
| | Other | 9.72 | |
| | Not Categorized | 5.64 | |
| RUCA 4 code class | Isolated | 2.30 | 0.97 |
| | Large Rural | 5.12 | |
| | Small Rural | 2.41 | |
| | Urban | 89.20 | |
| Study Entry CKD from eGFR lab | No | 83.17 | 0.00 |
| | Yes | 16.83 | |
| Study Entry CKD from Dx code | No | 97.06 | 0.00 |
| | Yes | 2.94 | |
| Study Entry CKD from Albuminuria/Proteinuria | No | 98.57 | 0.00 |
| | Yes | 1.43 | |
| Study Entry Diabetes Mellitus (DM) | No | 85.13 | 0.00 |
| | Yes | 14.87 | |
| Study Entry Pre-DM | No | 91.29 | 0.00 |
| | Yes | 8.71 | |
| Study Entry Hypertension (HTN) | No | 65.53 | 0.00 |
| | Yes | 34.47 | |
| Study Entry ACEIARB medication | No | 87.09 | 0.00 |
| | Yes | 12.91 | |
| Study Entry NSAID medication | No | 87.66 | 0.00 |
| | Yes | 12.34 | |
| Study Entry PPI medication | No | 91.32 | 0.00 |
| | Yes | 8.68 | |
| Year 1 ≥40% eGFR decline | No | 99.20 | 0.00 |
| | Yes | 0.80 | |
| Year 2 ≥40% eGFR decline | No | 98.96 | 0.00 |
| | Yes | 1.04 | |
| Year 3 ≥40% eGFR decline | No | 98.84 | 0.00 |
| | Yes | 1.16 | |
| Year 4 ≥40% eGFR decline | No | 98.79 | 0.00 |
| | Yes | 1.21 | |
| Year 5 ≥40% eGFR decline | No | 98.78 | 0.00 |
| | Yes | 1.22 | |
| Year 6 ≥40% eGFR decline | No | 98.82 | 0.00 |
| | Yes | 1.18 | |

Abbreviations: UCLA, University of California, Los Angeles Health; PHS, Providence Health System; RUCA (rural-urban commuting area, CKD (chronic kidney disease), eGFR (estimated glomerular filtration rate), DM (Diabetes Mellitus), ACEI (Angiotensin-converting enzyme), ARB (Angiotensin receptor blockers), NSAID (Non-steroidal anti-inflammatory drugs), PPI (Proton pump inhibitors).

**Table 3.** Continuous variables, average value, standard deviation, and percentage of missing values. The average age of participants was 56.78 ± 17.50 years, and the average number of years followed based on eGFR labs was 3.52 ± 2.74 years. Medication coverage is specified in days. Adopted from [19].

| Variable | Entry | | Year 1 | | Year 2 | | Year 3 | | Year 4 | | Year 5 | | Year 6 | |
|---|---|---|---|---|---|---|---|---|---|---|---|---|---|---|
| | Mean (SD) | Miss % | Mean (SD) | Miss % | Mean (SD) | Miss % | Mean (SD) | Miss % | Mean (SD) | Miss % | Mean (SD) | Miss % | Mean (SD) | Miss % |
| HbA1c (%) | 6.57 (1.66) | 79.53 | 6.52 (1.41) | 83.16 | 6.48 (1.41) | 84.32 | 6.47 (1.41) | 85.81 | 6.47 (1.40) | 87.31 | 6.47 (1.40) | 88.89 | 6.50 (1.41) | 90.59 |
| UACR (mg/g) | 108.98 (470.76) | 95.20 | 102.18 (439.80) | 94.57 | 95.23 (411.99) | 94.78 | 95.61 (407.72) | 95.20 | 93.80 (397.17) | 95.55 | 95.44 (402.22) | 95.55 | 101.34 (425.33) | 96.37 |
| UPCR (g/g) | 1.47 (2.72) | 99.51 | 1.26 (2.42) | 99.49 | 1.18 (2.31) | 99.60 | 1.17 (2.14) | 99.63 | 1.18 (2.22) | 99.66 | 1.15 (2.14) | 99.63 | 1.15 (2.24) | 99.71 |
| SBP (mmHg) | 128.19 (16.27) | 48.95 | 127.91 (15.33) | 45.58 | 128.21 (15.36) | 46.15 | 128.34 (15.24) | 47.41 | 128.52 (15.24) | 51.79 | 128.66 (15.24) | 57.20 | 128.79 (15.23) | 62.92 |
| DBP (mmHg) | 74.57 (10.36) | 49.03 | 75.40 (9.62) | 45.65 | 75.29 (9.61) | 46.21 | 75.45 (9.51) | 47.48 | 75.29 (9.48) | 51.85 | 75.16 (9.46) | 57.26 | 75.06 (9.44) | 62.96 |
| PP (mmHg) | 53.32 (13.31) | 48.91 | 52.48 (12.52) | 45.55 | 52.90 (12.70) | 46.11 | 52.87 (12.65) | 47.38 | 53.22 (12.71) | 51.76 | 53.48 (12.77) | 57.17 | 53.82 (12.77) | 62.89 |
| MAP (mmHg) | 92.80 (11.15) | 48.91 | 93.07 (10.42) | 45.55 | 93.28 (10.38) | 46.11 | 93.24 (10.30) | 47.38 | 93.20 (10.27) | 51.76 | 93.16 (10.25) | 57.17 | 93.14 (10.21) | 62.89 |
| ACEIARB coverage (days) | 7.10 (23.58) | 0.17 | 33.48 (99.25) | 0.36 | 31.94 (96.95) | 0.38 | 33.61 (98.11) | 0.53 | 33.69 (99.92) | 0.39 | 29.70 (94.69) | 0.28 | 25.42 (88.04) | 0.23 |
| NSAID coverage (days) | 5.32 (20.18) | 0.14 | 25.68 (84.80) | 0.31 | 25.16 (84.60) | 0.30 | 25.45 (84.75) | 0.33 | 24.97 (84.67) | 0.29 | 22.59 (81.14) | 0.24 | 19.38 (75.53) | 0.21 |
| PPI coverage (days) | 4.85 (19.39) | 0.16 | 20.96 (77.14) | 0.36 | 19.52 (74.92) | 0.37 | 20.57 (76.25) | 0.46 | 20.69 (77.87) | 0.35 | 18.45 (74.15) | 0.26 | 15.89 (69.06) | 0.22 |
| eGFR (ml/min/1.73 m$^2$) | 83.06 (23.97) | 0.12 | 80.03 (24.50) | 40.32 | 80.00 (23.82) | 48.04 | 79.57 (23.50) | 54.38 | 79.30 (23.22) | 59.90 | 78.94 (22.97) | 64.93 | 78.30 (22.77) | 70.03 |

Abbreviations: HbA1C (Hemoglobin A1c), UACR (urine albumin-creatinine ratio), UPCR (urine protein-creatinine ratio), SBP (systolic blood pressure), DBP (diastolic blood pressure), PP (pulse pressure), MAP (mean arterial pressure), eGFR (estimated glomerular filtration rate), DM (Diabetes Mellitus), ACEI (Angiotensin-converting enzyme), ARB (Angiotensin receptor blockers), NSAID (Non-steroidal anti-inflammatory drugs), PPI (Proton pump inhibitors).